\documentclass[runningheads]{llncs}
\usepackage[T1]{fontenc}
\usepackage{graphicx}

\usepackage{amsmath,amssymb}
\usepackage{booktabs}
\usepackage[table,xcdraw]{xcolor}
\begin{document}
\title{TSynD: Targeted Synthetic Data Generation for Enhanced Medical Image Classification}
\subtitle{Leveraging Epistemic Uncertainty to Improve Model Performance}
\titlerunning{Targeted Synthetic Data Generation for Medical Image Classification}
\author{Joshua Niemeijer\inst{3} \and
        Jan Ehrhardt\inst{1} \and
        Hristina Uzunova\inst{2} \and
        Heinz Handels\inst{1,2}}
\institute{
\inst{1}Institute of Medical Informatics, University of Lübeck, Germany \\
\inst{2}German Research Center for Artificial Intelligence, Lübeck, Germany \\
\inst{3}German Aerospace Center, Braunschweig, Germany \\ 
\email{Joshua.Niemeijer@dlr.de}
}
\maketitle              %
\begin{abstract}
The usage of medical image data for the training of large-scale machine learning approaches is particularly challenging due to its scarce availability and the costly generation of data annotations, typically requiring the engagement of medical professionals.
The rapid development of generative models allows towards tackling this problem by leveraging large amounts of realistic synthetically generated data for the training process.
However, randomly choosing synthetic samples, might not be an optimal strategy.

In this work, we investigate the targeted generation of synthetic training data, in order to improve the accuracy and robustness of image classification.
Therefore, our approach aims to guide the generative model to synthesize data with high epistemic uncertainty, since large measures of epistemic uncertainty indicate underrepresented data points in the training set.
During the image generation we feed images reconstructed by an auto encoder into the classifier and compute the mutual information over the class-probability distribution as a measure for uncertainty. 
We alter the feature space of the autoencoder through an optimization process with the objective of maximizing the classifier uncertainty on the decoded image. 
By training on such data we improve the performance and robustness against test time data augmentations and adversarial attacks on several classifications tasks.

\keywords{synthetic data generation \and generalization \and robustness}
\end{abstract}

\section{Introduction}
\label{sec::introduction}
Creating imaging datasets for training deep neural networks consist of three major steps:
data acquisition, data selection, and data labeling.
These steps are especially challenging in the domain of medical image processing. 
Data acquisition is often limited and data delivery is impaired by privacy regulations. Also, relevant image data might further be bound by the frequency of certain medical scenarios (e.g. rare diseases). Another main obstacle is the costly and time-intensive data labeling, which often requires medical professionals

In this work, we address these problems by utilizing generative models to extend the distribution of the given training data.
More specifically, we aim to create data points that represent missing parts of the relevant distribution.
Such data points are marked by a high epistemic uncertainty when processed by a discriminative model (i.e. a classifier network).
In this work, we present a novel approach called TSynD (\emph{\textbf{T}argeted \textbf{Syn}thetic \textbf{D}ata generation}): a method specifically designed to  steer the generation process in order to synthesize data points from the missing parts of the training distribution and utilize them during the training of downstream tasks (here: classifier). 
For data generation,
TSynD employs an autoencoder model that is able to reconstruct existing images of the training distribution.
The autoencoder consists of an encoder that transforms the image into the latent space and a decoder that reconstructs the input image from the latent space.
TSynD aims to optimize the latent space representations of the autoencoder in a way that the decoded images maximize the epistemic uncertainty in a given classifier. 
By further training the classifier on these images, we receive classification models that generalize well to unseen data, a feature that is especially important in medical image processing. 
We, therefore, show the performance of the TSynD method on several medical classification datasets. %
In order to simulate the smaller training datasets, typical for the medical image community, as well as recreate cases of out-of-distribution samples, this work primarily considers a low-data training setting.
We provide experiments to investigate the out-of-distribution performance through random test time augmentations and investigate the robustness to adversarial attacks. 
Further, the robustness of our approach is investigated visually by applying class activation explanation approaches and we are able to show that a classifier trained with TSynD utilizes more meaningful image information.

\section{Related Work}
\label{sec::related_work}
In our work, we present a novel method for training networks that generalize to out of distribution samples. 
We employ an adaptive data generation process that is based on generative models. 

Data augmentation is a commonly used way of extending the given training distributions mostly in an untargeted way.
As stated in Zhou et al.~\cite{Zhou2023Survey}, there are four different types of data augmentation:
Firstly there are image transformations, which consist of  e.g. random flipping, rotation or color augmentations. 
Secondly, model based augmentations, which, e.g., consist of random convolutions\cite{Xu2020RandConv} or other augmentation networks like style transfer networks \cite{borlino2021rethinking} or  learnable image generators \cite{zhou2020learning,Niemeijer_2024_WACV}. 
Thirdly, latent space augmentations, which directly augment the latent space distributions of the tasks model as in Zhou et al.~\cite{Zhou2021MixStyle}. 
Finally, there are approaches that utilize adversarial gradients. 

Adversarial gradient augmentation and, more specifically, task adversarial augmentations are the most similar category to our approach. 
The approaches of Sinha et al.~\cite{sinha2017certifying}, Volpi et al.~\cite{Volpi2018ADA} and Qiao et al.~\cite{qiao2020learning} utilize adversarial attacks by computing adversarial gradients  w.r.t. to the task network to alter the images that are used for training. 
The alternation is hereby done by optimizing the pixel values of the image as parameters themselves. 
Such  methods are often accused of introducing noise perturbations instead for larger image alternations e.g. representing domain shifts (Zhou et al.~\cite{Zhou2023Survey}). 

In contrast we optimize the latent space of a generative model as parameters of the image generation.
Since the latent space is a more abstract representation the idea is that altering in a target way results in larger more meaningful changes.
The work of Stutz et al.~\cite{Stutz2019CVPR} is therefore the most related to our approach. 
They employ a VAE-GAN model \cite{rosca2017variational} to represent the manifold and similar to us compute perturbation on the latent space to create adversarial examples. 
In contrast to us they do not maximize the uncertainty, but rather maximize the cross entropy loss, effectively changing the predicted label.
This introduces the need for constraints, in order to maintain the image  class.
Our approach is inspired by active learning \cite{BestPrAL2023} and puts the main focus on generating images that maximize the epistemic uncertainty of the given classifier. 
This brings the advantage of not requiring any additional constraints. 
The work of Li et al.~\cite{li2021simple} utilizes an autoencoder, also.  
Similar to us they compute perturbations on the latent space by e.g. utilizing random noise. 
In our work we utilize the randomly perturbed latent space as start for our optimization to increase the data diversity.

\section{Methods}
\label{sec::methods}
\begin{figure}[tbp]
  \centering
  \includegraphics[width=\linewidth]{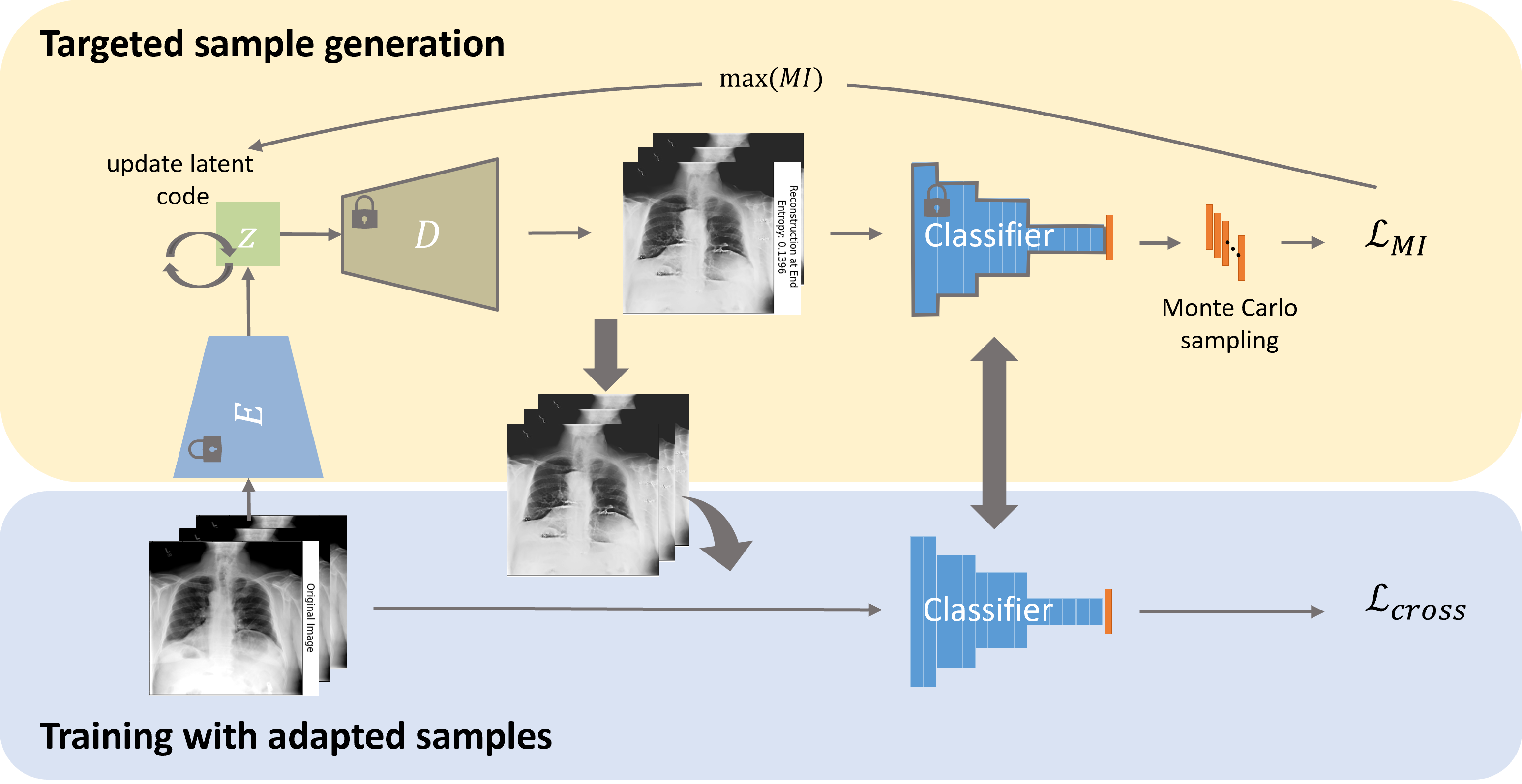}
  \caption{The overall framework of TSynD (\emph{\textbf{T}argeted \textbf{Syn}thetic \textbf{D}ata generation}) for the robust training of a classifier: During classifier training, the latent spatial representations of original images are optimized to maximize the classifier's epistemic uncertainty in the decoded images. The new images then serve as additional training data.}
  \label{fig:ADG_architecture}
\end{figure}

Given a labeled data set $\mathcal{D}=\{(x_n, y_n)\}_{n=1}^N$ $\left(x_n\in\mathcal{X}, y_n\in\mathcal{Y}\right)$, our approach aims to use the generative model in a targeted way to make a classification network $\mathcal{C}:\mathcal{X}\rightarrow\mathcal{Y}$ more robust to missing parts of the data distribution that are not included in the labeled set $\mathcal{D}$. The generative model, e.g. an autoencoder, consists of an encoding function $f_{\text{enc}}: \mathcal{X}\rightarrow\mathcal{Z}$ and a decoder $f_{\text{dec}}:\mathcal{Z}\rightarrow\mathcal{X}$, where $\mathcal{Z}$ is the latent space. It can be trained in an unsupervised way using a larger amount of unlabeled data from the domain $\mathcal{X}$.
Inspired by active learning strategies, we utilize the generative model to create images that maximize the epistemic uncertainty of our classification network $\mathcal{C}$. Samples yielding a high epistemic uncertainty represent missing parts of the learned distribution, and training on such samples can make the classification network more robust. 
Figure \ref{fig:ADG_architecture} shows an overview of our approach: starting by the encoded labeled images, the latent code $z$ is optimized to reconstruct new images that locally maximize the epistemic uncertainty  of the classifier $\mathcal{C}$. The newly generated samples are now used together with the labeled images for the training of the classifier.

\subsection{Estimation of the epistemic uncertainty}
Given the classifier $\mathcal{C}$ with model parameters $\theta$, the predictive class probability distribution for a decoded image $\hat{x}=f_{\text{dec}}(z)$ with latent code $z$ is computed by
$$
p(y | \hat{x}, \theta) = p(y | z, \theta) = \sigma(\mathcal{C}(f_{\text{dec}}(z); \theta)) ,
$$
where the function $\sigma(\cdot)$ transfers the classifier logit outputs into probabilities. 
Here, $\sigma(\cdot)$ is the softmax function in case of a multilabel classification or a sigmoid function in case of a binary classification. 
The primary objective is to guide the reconstruction process $\hat{x}=f_{\text{dec}}(z)$ in a manner that the resulting sample $\hat{x}$ contributes meaningfully to the training of the classifier $\mathcal{C}$. 
This guidance involves modifying a latent variable $z\in\mathcal{Z}$ of the autoencoder with the aim of generating samples with a high epistemic uncertainty in the classifier $\mathcal{C}$.

The uncertainty of the predictive distribution is defined by the entropy
$$
\mathbf{U}_H(z) = H(p(y|z,\theta)) = -\sum_{y\in\mathcal{Y}} \hat{p}(y | z,\theta) \log(\hat{p}(y | z, \theta)),
$$
however, the epistemic uncertainty associated with a data sample $\hat{x}=f_{\text{dec}}(z)$ stems from uncertainty in model parameters. This can be quantified by the expected change in entropy of the model parameter posterior distribution, expressed by the conditional mutual information~\cite{linander2023looking}:
$$
\mathbf{U}_{MI}(z)= MI(z;\theta) = H(\mathbb{E}_\theta(p(y|z, \theta)))- \mathbb{E}_\theta(H(p(y|z, \theta))),
$$
where the expectation is computed over Monte Carlo Dropouts ~\cite{Kirsch2019BatchBALD}. The mutual information is considered to be a better measure for the epistemic uncertainty \cite{Kirsch2019BatchBALD}. To keep the additional computational effort low we only iterate
over the last layers of $\mathcal{C}$ with $K$ dropout masks to compute samples of $p(y|z, \theta^{k})$, $k=1\ldots K$.

\subsection{Targeted Synthetic Data Generation}
An optimization-based approach is used to find latent codes $z$ that locally maximize the given measure for uncertainty $\mathbf{U}(z)$. %
As shown in Fig. \ref{fig:ADG_architecture}, starting with the latent code $z_n=f_{\text{enc}}(x_n)$ of a random image of the training distribution, we search a local maximum
\begin{equation}\label{eq:maxU}
z^*=\arg\max_z \ \mathbf{U}(z).    
\end{equation}
Since $\mathcal{C}$ and $f_{\text{dec}}$ both are differentiable, $\mathbf{U}(z)$ can be maximized by standard backpropagation. The resulting sample $\hat{x} = f_{\text{dec}}(z^*)$ is added to the training set, assuming that it belongs to the same class as $x_n$, but lies in missing parts of the learned data distribution, as indicated by the high uncertainty.

\subsubsection{Latent space noise.}
Apart form the uncertainty that a sample yields, the diversity w.r.t. the training distribution is crucial. 
To introduce further varieties into the reconstructed image we generate samples by adding uniform noise to latent codes $z_n=f_{\text{enc}}(x_n)$:
$$
\hat{x} = f_{\text{dec}}(z_n + \epsilon),\quad \epsilon\sim N(0,\sigma\mathbf{I}).
$$
The resulting image $\hat{x}$ is an alternative representation of $x_n$ and hence increases the diversity in the dataset. Latent space noise can be used as a stand-alone augmentation or as an initial augmentation before optimization according to Eq.~\ref{eq:maxU} to generate even more diverse samples. 
\subsubsection{The training process.}
In each training iteration, the batch is divided into two halves:
The first half consists of original image-label pairs $\{(x_n, y_n)\}_{n=1}^{\frac{B}{2}}$ and the second half consists of the optimized reconstructions $\hat{x}_n=f_{\text{dec}}(z^*_n)$ resulting from Eq. (\ref{eq:maxU}), along with their corresponding labels $\{(\hat{x}_n, y_n)\}_{n=1}^{\frac{B}{2}}$. 
Since the maximization of the epistemic uncertainty depends on the current state of the classifier $\mathcal{C}$ we need to redo the generation process of $\hat{x}_n$ after each training iteration. 
This also prevents the so called mode collapse problem that would occur if would run the image generation only once.
The resulting images would be similar, since similar images are likely to maximize the uncertainty of the given classifier. 
However since we retrain and generate in a alternating way, the network is updated and the generation process yields new alternations. 
\subsubsection{Optimizing latent codes vs. pixel values as parameters.}
Optimizing the pixel values like in \cite{sinha2017certifying,Volpi2018ADA,qiao2020learning} is likely results in salt and peper noise.
Altering abstract representations gives us more substantial alternations, since each element of $z\in\mathcal{Z}$ represents larger receptive fields in the image. 
Additionally the autoencoder is learned on the distribution of relevant images, already. The reconstruction process hence is already constrained w.r.t. to this distribution. 
Constraints that need to be introduces when optimizing the image pixels directly like in the approaches of \cite{sinha2017certifying,Volpi2018ADA,qiao2020learning} are not needed. 
It is however important to maximize the epistemic uncertainty (the model uncertainty) and not the aleatoric uncertainty. The maximizing the latter would result in ambiguous data, since the aleatoric uncertainty represents the data uncertainty.

\section{Experiments}
\label{sec::experiments}
\begin{table}[tb]
\begin{center}
\resizebox{\textwidth}{!}{
\begin{tabular}{@{}clcclcclcclcclcclcc@{}}
\toprule
\multicolumn{1}{c|}{\cellcolor[HTML]{ECF4FF}} &  \multicolumn{1}{c|}{} & \multicolumn{2}{c|}{\cellcolor[HTML]{ECF4FF}\textbf{BreastMNIST}}                                      &  \multicolumn{1}{c|}{} & \multicolumn{2}{c|}{\cellcolor[HTML]{ECF4FF}\textbf{DermaMNIST}}                                       &  \multicolumn{1}{c|}{} & \multicolumn{2}{c|}{\cellcolor[HTML]{ECF4FF}\textbf{OCTMNIST}}                                                          &  \multicolumn{1}{c|}{} & \multicolumn{2}{c|}{\cellcolor[HTML]{ECF4FF}\textbf{OrganaMNIST}}                                      &  \multicolumn{1}{c|}{} & \multicolumn{2}{c|}{\cellcolor[HTML]{ECF4FF}\textbf{OrgansMNIST}}                                      &  \multicolumn{1}{c|}{} & \multicolumn{2}{c}{\cellcolor[HTML]{ECF4FF}\textbf{PathMNIST}}                    \\
\multicolumn{1}{c|}{\cellcolor[HTML]{ECF4FF}} &  \multicolumn{1}{c|}{} & \cellcolor[HTML]{ECF4FF}\textbf{1\%}    & \multicolumn{1}{c|}{\cellcolor[HTML]{ECF4FF}\textbf{10\%}}   &  \multicolumn{1}{c|}{} & \cellcolor[HTML]{ECF4FF}\textbf{1\%}    & \multicolumn{1}{c|}{\cellcolor[HTML]{ECF4FF}\textbf{10\%}}   &  \multicolumn{1}{c|}{} & \multicolumn{1}{c}{\cellcolor[HTML]{ECF4FF}\textbf{1\%}} & \multicolumn{1}{c|}{\cellcolor[HTML]{ECF4FF}\textbf{10\%}}   &  \multicolumn{1}{c|}{} & \cellcolor[HTML]{ECF4FF}\textbf{1\%}    & \multicolumn{1}{c|}{\cellcolor[HTML]{ECF4FF}\textbf{10\%}}   &  \multicolumn{1}{c|}{} & \cellcolor[HTML]{ECF4FF}\textbf{1\%}    & \multicolumn{1}{c|}{\cellcolor[HTML]{ECF4FF}\textbf{10\%}}   &  \multicolumn{1}{c|}{} & \cellcolor[HTML]{ECF4FF}\textbf{1\%}    & \cellcolor[HTML]{ECF4FF}\textbf{10\%}   \\ \midrule
\multicolumn{1}{c|}{Baseline}                 &  \multicolumn{1}{c|}{} & 70.7                                    & \multicolumn{1}{c|}{75.2}                                    &  \multicolumn{1}{c|}{} & 66.8                                    & \multicolumn{1}{c|}{65.2}                                    &  \multicolumn{1}{c|}{} & 59.9                                                     &  \multicolumn{1}{c|}{67.5}                                    &  \multicolumn{1}{c|}{} & 71.9                                    & \multicolumn{1}{c|}{89.3}                                    &  \multicolumn{1}{c|}{} & 49.6                                    & \multicolumn{1}{c|}{67.8}                                    &  \multicolumn{1}{c|}{} & 67.4                                    & 77.2                                    \\
\multicolumn{1}{c|}{Noise}                    &  \multicolumn{1}{c|}{} & 72.0                                    & \multicolumn{1}{c|}{\textbf{78.2}}                           &  \multicolumn{1}{c|}{} & 66.7                                    & \multicolumn{1}{c|}{65.8}                                    &  \multicolumn{1}{c|}{} & 61.0                                                     &  \multicolumn{1}{c|}{\textbf{ 71.7 }} &  \multicolumn{1}{c|}{} & 73.8                                    & \multicolumn{1}{c|}{87.8}                                    &  \multicolumn{1}{c|}{} & 52.9                                    & \multicolumn{1}{c|}{68.6}                                    &  \multicolumn{1}{c|}{} & 64.7                                    & \textbf{ 83.5 } \\
\multicolumn{1}{c|}{TSynD}                    &  \multicolumn{1}{c|}{} & \textbf{73.3}                           & \multicolumn{1}{c|}{77.8}                                    &  \multicolumn{1}{c|}{} & \textbf{ 66.9 } & \multicolumn{1}{c|}{\textbf{ 66.7 }} &  \multicolumn{1}{c|}{} & \textbf{ 61.4 }                  &  \multicolumn{1}{c|}{66.7}                                    &  \multicolumn{1}{c|}{} & \textbf{ 77.2 } & \multicolumn{1}{c|}{\textbf{ 89.4 }} &  \multicolumn{1}{c|}{} & \textbf{ 54.2 } & \multicolumn{1}{c|}{\textbf{ 71.4 }} &  \multicolumn{1}{c|}{} & \textbf{ 73.1 } & 78.5                                    \\ \midrule
\multicolumn{19}{c}{\cellcolor[HTML]{ECF4FF}\textbf{Gaussian Noise Augmentation during Test}}                                                                                                                                                                                                                                                                                                                                                                                                                                                                                                                                                                                                                                                                                                                                                   \\ \midrule
\multicolumn{1}{c|}{Baseline}                 &  \multicolumn{1}{c|}{} & 62.6                                    & \multicolumn{1}{c|}{\textbf{ 73.1 }} &  \multicolumn{1}{c|}{} & 66.8                                    & \multicolumn{1}{c|}{65.1}                                    &  \multicolumn{1}{c|}{} & 24.9                                                     &  \multicolumn{1}{c|}{29.9}                                    &  \multicolumn{1}{c|}{} & 44.5                                    & \multicolumn{1}{c|}{78.9}                                    &  \multicolumn{1}{c|}{} & 37.1                                    & \multicolumn{1}{c|}{52.6}                                    &  \multicolumn{1}{c|}{} & 12.6                                    & 10.6                                    \\
\multicolumn{1}{c|}{Noise}                    &  \multicolumn{1}{c|}{} & \textbf{73.5}                           & \multicolumn{1}{c|}{\textbf{ 73.1 }} &  \multicolumn{1}{c|}{} & 66.7                                    & \multicolumn{1}{c|}{65.7}                                    &  \multicolumn{1}{c|}{} & 24.5                                                     &  \multicolumn{1}{c|}{34.9}                                    &  \multicolumn{1}{c|}{} & 44.1                                    & \multicolumn{1}{c|}{65.3}                                    &  \multicolumn{1}{c|}{} & 37.3                                    & \multicolumn{1}{c|}{52.4}                                    &  \multicolumn{1}{c|}{} & 13.5                                    & 11.5                                    \\
\multicolumn{1}{c|}{TSynD}                    &  \multicolumn{1}{c|}{} & 73.3                                    & \multicolumn{1}{c|}{\textbf{ 73.1 }} &  \multicolumn{1}{c|}{} & \textbf{ 66.9 } & \multicolumn{1}{c|}{\textbf{66.7}}                           &  \multicolumn{1}{c|}{} & \textbf{ 28.7 }                  &  \multicolumn{1}{c|}{\textbf{ 36.4 }} &  \multicolumn{1}{c|}{} & \textbf{ 63.5 } & \multicolumn{1}{c|}{\textbf{ 85.1 }} &  \multicolumn{1}{c|}{} & \textbf{ 45.6 } & \multicolumn{1}{c|}{\textbf{ 66.4 }} &  \multicolumn{1}{c|}{} & \textbf{ 28.2 } & \textbf{ 12.8 } \\ \midrule
\multicolumn{19}{c}{\cellcolor[HTML]{ECF4FF}\textbf{Adversarial Attacks during Test}}                                                                                                                                                                                                                                                                                                                                                                                                                                                                                                                                                                                                                                                                                                                                                           \\ \midrule
\multicolumn{1}{c|}{Baseline}                 &  \multicolumn{1}{c|}{} & 65.6                                    & \multicolumn{1}{c|}{7.1}                                     &  \multicolumn{1}{c|}{} & 66.4                                    & \multicolumn{1}{c|}{48.8}                                    &  \multicolumn{1}{c|}{} & 5.3                                                      &  \multicolumn{1}{c|}{3.3}                                     &  \multicolumn{1}{c|}{} & 34.4                                    & \multicolumn{1}{c|}{68.4}                                    &  \multicolumn{1}{c|}{} & 13.8                                    & \multicolumn{1}{c|}{25.6}                                    &  \multicolumn{1}{c|}{} & 28.5                                    & 21.1                                    \\
\multicolumn{1}{c|}{Noise}                    &  \multicolumn{1}{c|}{} & 68.6                                    & \multicolumn{1}{c|}{21.6}                                    &  \multicolumn{1}{c|}{} & \textbf{ 66.7 } & \multicolumn{1}{c|}{53.0}                                    &  \multicolumn{1}{c|}{} & 8.1                                                      &  \multicolumn{1}{c|}{4.1}                                     &  \multicolumn{1}{c|}{} & 39.2                                    & \multicolumn{1}{c|}{71.6}                                    &  \multicolumn{1}{c|}{} & 25.1                                    & \multicolumn{1}{c|}{25.6}                                    &  \multicolumn{1}{c|}{} & 31.7                                    & 26.1                                    \\
\multicolumn{1}{c|}{TSynD}                    &  \multicolumn{1}{c|}{} & \textbf{ 71.4 } & \multicolumn{1}{c|}{\textbf{ 28.2 }} &  \multicolumn{1}{c|}{} & \textbf{ 66.7 } & \multicolumn{1}{c|}{\textbf{ 64.1 }} &  \multicolumn{1}{c|}{} & \textbf{ 12.8 }                  &  \multicolumn{1}{c|}{\textbf{ 42.8 }} &  \multicolumn{1}{c|}{} & \textbf{ 53.5 } & \multicolumn{1}{c|}{\textbf{ 83.9 }} &  \multicolumn{1}{c|}{} & \textbf{ 27.1 } & \multicolumn{1}{c|}{\textbf{ 51.3 }} &  \multicolumn{1}{c|}{} & \textbf{ 43.5 } & \textbf{ 47.8 } \\ \bottomrule
\end{tabular}}
\end{center}
\caption{Accuracy results of different MedMNIST datasets with a subsampling of the training dataset to $1\%$ and $10\%$. The results are reported for the respective test set of the datasets and two augmented versions of the tests sets (Gaussian Noise and adversarial attacks).}
\label{tab:EvaluationResults}
\end{table}

Our experiments aim to show the effect of TSynD on the generalization performance and robustness of classification networks. 
Since the test and validation sets of available datasets are often drawn from similar distributions as the training distribution, the generalization of networks is hard to measure. 
For that reason we introduce a sub-sampling of the training dataset to 1\% and 10\% of the respective datasets. 
This introduces a sampling bias and makes it more likely that the test and validation distributions contain out of distribution data.
This also mirrors the common scenario in medical data where training datasets are often small.
Our experiments concentrate on two main questions: 1) Does the proposed TSynD  improve classification results when training in a low-data setting?
2) Is the training using the proposed approach more robust, e.g., against random test data augmentations and test time adversarial attacks? To investigate 1), we train and evaluate using three different settings: 
baseline classifier without any additional training time augmentations; 
augmentation through random latent space noise during the training (see section \ref{sec::methods}); 
and training using TSynD. 
For research question 2), the three previously trained settings are used and the tested in three scenarios: 
no test data augmentation;  Gaussian noise with $\sigma =0.2$ added to the test data; and the test data is altered using adversarial attacks as described in \cite{goodfellow2014explaining}.

The datasets used in our experiments are MedMNIST v2 \cite{Yang2021MedMNISTv2} datasets and the Chest-Xray \cite{Wang2017ChestXRay} dataset for classification, since they are openly available and suitable for establishing a baseline. 
We utilized the commonly used ResNet-18 \cite{he2015deep} and DenseNet \cite{huang2018densely} as classifiers, and a state-of-the-art autoencoder VQ-VAE~\cite{oord2018neural,esser2021taming} trained unsupervised on the full training set as the generative model. In each experiment, the classifier was trained for 100 epochs and the model with the best validation performance selected. Training was repeated three times and averaged values are reported.
Our TSynD process is influenced by the learning rate (chosen as $0.1$) and the number of iterations (either $100$ or $50$) for the optimizer to maximize the epistemic uncertainty. 
The noise factor that is added to the feature space is chosen empirically (either $0.1$ or $1.0$ in our experiments).

\subsection{Classification and robustness results}
Table \ref{tab:EvaluationResults} shows the classification results across different MedMNIST datasets using a ResNet-18 model to compare baseline training without augmentation, augmentation with latent space noise (Noise) and TSynD.
The TSynD models improve over the baseline model and even over the Noise models on the standard test sets in almost all low data scenarios that were tested.
This shows that TSynD is an effective method for training models that generalize well in such low data settings. It further shows the advantage of the targeted optimization-based generation of new samples compared to a random sampling. 
When we apply Gaussian noise to the test set, or introduce adversarial attacks we can observe that the TSynD models are always better than the baseline models and even improve over the noise models, as well. 
This indicates that the samples that were generated by TSynD made the resulting model more robust against these out of distribution samples. 

\subsection{Uncertainty Maximization}

\begin{table}[tb]
\begin{center}
\begin{tabular}{@{}c|l|*{2}{p{1.5cm}|}|l|*{2}{p{1.5cm}|}|l|*{2}{p{1.5cm}}@{}}
\toprule
\rowcolor[HTML]{ECF4FF} 
        &  & \multicolumn{2}{c}{\cellcolor[HTML]{ECF4FF}\textbf{OrgansMNIST}} &  & \multicolumn{2}{c}{\cellcolor[HTML]{ECF4FF}\textbf{Chest-XRay}} &  & \multicolumn{2}{c}{\cellcolor[HTML]{ECF4FF}\textbf{OCTMNIST}} \\
\rowcolor[HTML]{ECF4FF} 
        &  & \multicolumn{1}{c}{1\%}                         & \multicolumn{1}{c}{10\%}                       &  & \multicolumn{1}{c}{1\%}                       & \multicolumn{1}{c}{10\%}                      &  & \multicolumn{1}{c}{1\%}                       & \multicolumn{1}{c}{10\%}                     \\ \midrule
Entropy &  & \multicolumn{1}{c}{73.5}                        & \multicolumn{1}{c}{86.0}                       &  & \multicolumn{1}{c}{60.8}                       & \multicolumn{1}{c}{67.9}                       &  & \multicolumn{1}{c}{79.5}                      & \multicolumn{1}{c}{82.9}                      \\
MI      &  & \multicolumn{1}{c}{68.1}                       & \multicolumn{1}{c}{84.1}                       &  & \multicolumn{1}{c}{61.5}                       & \multicolumn{1}{c}{68.0}                       &  & \multicolumn{1}{c}{81.6}                      & \multicolumn{1}{c}{89.6}                      \\ \bottomrule
\end{tabular}%
\label{tab:MIvsEntropy}
\end{center}
\caption{Comparison between ADG maximizing mutual information (MI) and Entropy on the validation sets of the respective datasets.}
\end{table}

\label{sec::ablation}
Table \ref{tab:MIvsEntropy} presents an ablation study w.r.t. to the uncertainty that is maximized during the image generation.
In section \ref{sec::methods} we introduce the entropy and the mutual information ($MI$) as measures for the uncertainty. 
We can see that the $MI$ performs better than the entropy. 
This also aligns with the theory, since the entropy often is rather viewed as a measure for the aleatoric uncertainty and the $MI$ is viewed as a measure for the epistemic uncertainty. 
However the difference between maximizing the $MI$ and the entropy are not large, indicating that the entropy is not a strict measure for the aleatoric uncertainty and the $MI$ is not a strict measure for the epistemic uncertainty.

\subsection{Qualitative Robustness Evaluation}
\begin{figure}[tb]
  \centering
  \includegraphics[width=1\linewidth]{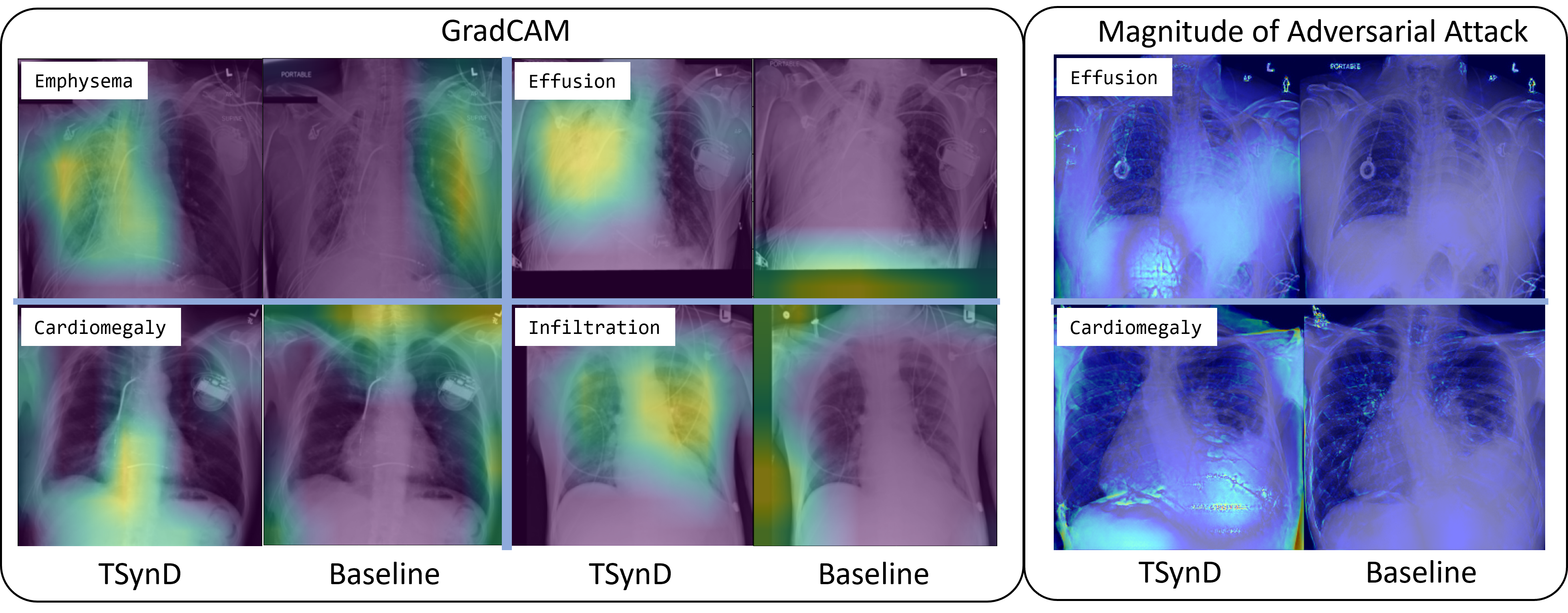}
  \caption{Left: EigenGradCAM maps of the baseline classifier and classifier trained with TSynD. Right: Perturbation of images to minimize the probability of the given class. Depicted is the difference of the images at the start and end of the minimization.}%
  \label{fig:explainability}
\end{figure}
We trained a classifier on the Chest-Xray \cite{Wang2017ChestXRay} dataset with and without TSynD. 
On average, we obtained an AUC improvement of about $1\%$ using TSynD on the validation set (both on the $1\%$ and $10\%$ subsampling of the training dataset). 
In this experiment, however, we do not concentrate on performance gain, moreover, we investigate the robustness of the proposed training mechanism. We explore the reasoning process of the classifier, by applying a commonly used explanation approach -- EigenGradCAM \cite{muhammad2020eigen}. The results can be seen on the left hand side in figure \ref{fig:explainability}. 
It can be observed, that the classifier trained using  TSynD utilizes more relevant regions of the image than the baseline classifier trained without TSynD.
We, additionally, employed our synthetic data generation process to create adversarial examples by minimizing class probabilities instead of maximizing the classifier uncertainty. 
The magnitude of difference between the original reconstruction and the optimized adversarial image can be seen on the right hand side of figure \ref{fig:explainability}. 
We  can observe that in order to minimize the probability for the classifier trained with TSynD, much larger and more relevant image regions must be altered, further indicating the increased robustness introduced by TSynD.

\section{Conclusion}
\label{sec::conclusion}
In this work we have shown how to utilize generative models to create synthetic data that is exploring unknown and relevant parts of the training distribution. 
We hence take a first step towards replacing the acquisition of large real world data distributions to select important data points with a more targeted generation process.
We have shown that training on this synthetic data yields a model that generalizes better to out of distribution samples and is more robust against adversarial attacks. 

In the current state our generation method only augments given samples. 
This in not ideal from a distribution diversity standpoint. 
As a future direction we want to extend the method to generate new samples that yield a high epistemic uncertainty and are therefore relevant for the training process.

\bibliography{report} %
\bibliographystyle{splncs04}

\end{document}